\def\year{2021}\relax
\documentclass[letterpaper]{article} 
\usepackage{aaai21}  
\usepackage{times}  
\usepackage{helvet} 
\usepackage{courier}  
\usepackage[hyphens]{url}  
\usepackage{graphicx} 
\usepackage{graphicx}  
\usepackage{natbib}  
\usepackage{caption} 
\usepackage{amsmath}
\usepackage{bm}
\nocopyright
\title{Empirical Evaluation of Deep Learning Model Compression\\Techniques on the WaveNet Vocoder}
\author{

    Sam Davis,
    Giuseppe Coccia,
    Sam Gooch,
    Julian Mack
    \\
}
\affiliations{

    Myrtle.ai \\

    Cambridge, UK \\
    sam@myrtle.ai, 
    giuseppe@myrtle.ai, 
    samg@myrtle.ai,
    julian@myrtle.ai

}

\begin{document}

\maketitle

\begin{abstract}
WaveNet is a state-of-the-art text-to-speech vocoder that remains challenging to deploy due to its autoregressive loop. In this work we focus on ways to accelerate the original WaveNet architecture directly, as opposed to modifying the architecture, such that the model can be deployed as part of a scalable text-to-speech system. We survey a wide variety of model compression techniques that are amenable to deployment on a range of hardware platforms. In particular, we compare different model sparsity methods and levels, and seven widely used precisions as targets for quantization; and are able to achieve models with a compression ratio of up to 13.84 without loss in audio fidelity compared to a dense, single-precision floating-point baseline. All techniques are implemented using existing open source deep learning frameworks and libraries to encourage their wider adoption.
\end{abstract}

\section{Introduction}

The widespread adoption of personal assistants has been powered, in large part, by advances in the field of Text-to-Speech~(TTS). Many state-of-the-art TTS systems contain a model, referred to as a vocoder, that takes as input audio features derived from a piece of text and outputs synthesized speech. WaveNet \cite{oord2016wavenet} is a state-of-the art vocoder that is capable of producing synthesized speech with near-human-level quality \cite{shen2018natural}. The key to the model's quality is its autoregressive loop but this property makes the model exceptionally challenging to deploy in applications that require real-time output or need to efficiently scale to millions of users since na\"ive implementations may take tens of minutes to generate ten seconds of speech.

As a result, TTS research has focused on finding alternative vocoder architectures such as Parallel-WaveNet \cite{oord2018parallel}, WaveRNN \cite{kalchbrenner2018efficient}, ClariNet \cite{ping2018clarinet} and WaveGlow \cite{prenger2019waveglow} that achieve higher performance when deployed on existing hardware. There is a degree of ambiguity as to the highest quality vocoder as audio quality evaluation is subjective but all authors agree that WaveNet produces at least as good if not higher quality audio than the more recent approaches \cite{kim2018flowavenet,oord2018parallel,prenger2019waveglow,tian2020featherwave,hsu2020wg}.

In this paper, rather than changing the WaveNet architecture to improve inference performance we instead keep it fixed and explore a range of model compression techniques that can yield greater inference performance. Crucially, the techniques we explore are all available in existing deep learning frameworks and are deployable to a wide range of current and future CPUs and neural network accelerators. Finally, motivated by the desire to maintain WaveNet's quality, we evaluate the impact that these compression techniques have on the perceived fidelity of the synthesized speech.

We examine two main categories of model compression---sparsity and quantization---and explore both their independent and combined impact on model quality. For sparsity we consider iterative and one-shot magnitude-based neural network pruning; and for quantization we explore the INT8, bfloat16, half-precision floating-point with both 16-bit and 32-bit accumulation (FP16.16, FP16.32), 16-bit block floating-point (BFP16), TensorFloat32 (TF32) and single-precision floating-point (FP32) formats. While there has been some work in vocoder pruning of WaveRNN and its variants \cite{kalchbrenner2018efficient,valin2019lpcnet,tian2020featherwave}, to our knowledge, this is the first paper in which WaveNet pruning results are provided. Additionally, to our knowledge, no other authors compare as wide a range of precisions, nor look at the interactions between pruning and quantization for the WaveNet model.

We include samples generated by the models presented in this work and will release our code \footnote{myrtlesoftware.github.io/wavenet-paper/}.

\section{Related Work}\label{sec:related_work}

Numerous attempts have been made to improve vocoder inference performance. The WaveRNN \cite{kalchbrenner2018efficient} authors note that the time for vocoder inference, $T(\bm{u})$, for a target audio sequence $\bm{u}$ can be decomposed into computation time $c_i$ and kernel launch overhead $d_i$ for each of the $N$ operations (layers) of the model:

\begin{equation}\label{eq:wn_optimization}
    T(\bm{u}) = |\bm{u}|\sum^N_{i=1}(c_i + d_i)
\end{equation}

Attempts to optimize a vocoder for deployment aim to reduce at least one of $\{|\bm{u}|, N, c_i, d_i\}$. Many approaches including Parallel-WaveNet, ClariNet, WaveGlow and WaveFlow \cite{ping2019waveflow} remove the autoregressive component and hence reduce $|\bm{u}|$ so that many or all of the samples can be generated in parallel. Others, including WaveRNN and its variants LPCNet \cite{valin2019lpcnet} and FeatherWave \cite{tian2020featherwave}, keep the autoregressive component but make alterations to the architecture to decrease the product of $N$ and $c_i$. There have also been efforts that focus on reducing $d_i$ by exploiting techniques such as persistent kernels that only launch the kernel once per sequence \cite{pharris_2018}. In this work we explore reducing $c_i$ without altering the WaveNet architecture by utilising model compression to give greater possibilities in the types of models that can be deployed.

\subsection{Sparsity}

Sparse models offer two potential benefits over their dense counterparts:

\begin{enumerate}
    \item The amount of computation can be reduced since, for example, multiplications by zero need not be performed.
    \item Memory bandwidth requirements can be reduced as it is possible to achieve higher compression ratios with sparse matrices.
\end{enumerate}

The first of these reduces $c_i$ but in order to realise either of these benefits, hardware support is usually required. Depending on the type of support, different hardware platforms are amenable to different types of sparsity. At one end of the spectrum, some authors use channel pruning, in which entire convolutional channels are set to zero \cite{he2017channel}. It is comparatively easy to realise the inference-time performance benefits of channel sparsity but this approach produces a significant degradation in audio quality for WaveNet \cite{hussain2020fastwave}. Channel sparsity is a special case of block sparsity where for a 2D matrix, blocks of size $n \times m$ are enforced to be either all-dense or all-sparse \cite{narang2017block}. At the other end of the spectrum, sparsity can also be unstructured meaning there are no constraints on the sparsity pattern; this typically results in the smallest quality degradation but is also the most challenging sparsity pattern to deploy efficiently. A hybrid approach is to employ balanced sparsity \cite{yao2019balanced,cao2019efficient} where each block is independently pruned to the target sparsity percentage but within a block the sparsity is unstructured.

The other principal axes on which neural network sparsity approaches can differ relate to the method of obtaining the sparse network. One way to do this is by utilizing magnitude-based pruning, an approach in which the parameters closest to zero are pruned \cite{han2015learning}. Magnitude-based pruning can be broadly classified as either one-shot in which a trained network is pruned to the desired sparsity in a single step \cite{han2015learning} or iterative where the level of sparsity increases more gradually over the training process \cite{zhu2017prune}. However, the division between the two is not always clear-cut. For example, the FeatherWave authors propose a two-stage sparse pruning schedule (TSSP) which combines a one-shot jump to 50\% sparsity followed by an iterative pruning stage. 

The WaveRNN, LPCNet and FeatherWave models utilise high levels of sparsity to reduce inference time. The WaveRNN authors use iterative pruning to achieve sparsity as high as 96\%. They investigate a range of block sparsity patterns including $1\text{x} 1$ (unstructured), $4\text{x} 4$ and $16\text{x} 1$ and find that the latter is the most performant as it more closely mirrors the layout in physical memory. The LPCNet and FeatherWave authors both use 90\% sparse networks with a $16\text{x} 1$ pattern although the latter uses TSSP as discussed above instead of the purely iterative approach.

\subsection{Quantization}\label{sec:related_work_quantization}

Quantized models also reduce $c_i$ as the operations are now performed at a numerical format in which the operations are less computationally expensive. This approach has been applied to a wide variety of models including BERT \cite{wu2020integer}, ResNet and GNMT \cite{zafrir2019q8bert}, and sees adoption in widely recognised machine learning benchmarks \cite{reddi2019mlperf}.

The quantization process includes one or both of:

\begin{enumerate}
    \item Reducing the number of bits of the datatype. e.g. use 8 bits instead of 32 bits.
    \item Using a less expensive format. e.g. use integer instead of floating-point.
\end{enumerate}

A simple scheme is to perform all multiplications in the FP16 data format as this is already widely supported on a variety of hardware devices. The results are accumulated in either FP16 or FP32; this distinction matters for the range of representable values and for what precision any activation functions are later performed in. We represent these choices as FP16.16 and FP16.32 to represent using FP16 for multiplies and either FP16 or FP32 for accumulations respectively.

Quantizing to integers is another popular choice for quantization. When quantizing from floating point to integer, it is necessary to use a quantization scheme in which there is no quantization error for the value $0$ as these parameters will have an outsized impact on model performance \cite{jacob2018quantization} especially when quantizing sparse matrices. Running inference at INTX (most often INT8) is widely used for deployment including for models from the domains of Machine Translation \cite{wu2016google}, Automatic Speech Recognition \cite{he2019streaming}, Computer Vision \cite{wu2018training} and NLP embeddings \cite{zafrir2019q8bert}.

However, formats other than INTX and the basic FP16 are becoming more widely used as hardware vendors accommodate them. For example, BrainFloat16 is supported by Google TPUs\footnote{cloud.google.com/tpu/docs/bfloat16} and Intel engineers have demonstrated a 1.5$\times$ speedup using this format for both Parallel-WaveNet and FeatherWave on Xeon CPUs\footnote{intel.com/content/www/us/en/artificial-intelligence/posts/intel-xeon-text-to-speech.html}. BlockFloat16 is supported on Intel's Stratix 10 NX FPGA\footnote{intel.com/content/www/us/en/products/programmable/fpga/ stratix-10/nx.html} and NVIDIA's Ampere series of GPUs will support the TensorFloat32 format. In the FPGA space, there is work using bespoke formats; for example, \cite{hussain2020fastwave} achieved a 2$\times$ speedup for WaveNet inference quantizing from floating point to fixed point while keeping the range and precision of their 27-bit data type constant. 

In many cases, it is possible to perform post-training quantization (PTQ), in which full-precision weights are quantized to the desired precision after training is completed, with minimal loss in model quality. However, when the quality degradation is large, it becomes necessary to perform Quantization Aware Training (QAT) \cite{jacob2018quantization}. In QAT, the quantization operation for the target precision is simulated during training so that the trained FP32 weights learn to remove the noise injected by the quantization.

\begin{table*}[t]
    \centering
    \begin{tabular}{l|l|r|r|r|r}
        Layer & Type & \multicolumn{2}{c|}{\# Parameters} & \multicolumn{2}{c}{GOP/second audio} \\
        & & Per-layer & Total & Per-layer & Total \\
        \hline
        \hline
        \textit{Pre-processing Layers} & & \multicolumn{2}{r|}{} & \multicolumn{2}{c}{} \\
        Embedding & Embedding & \multicolumn{2}{r|}{30,720} & \multicolumn{2}{r}{-} \\
        Feature Upsample & ConvTranspose1d & \multicolumn{2}{r|}{5,120,080} & \multicolumn{2}{r}{0.82} \\
        \hline
        \textit{Repeated Layers} & & & & & \\
        Dilation & Dilated Conv1d & 57,840 & 925,440 & 1.84 & 29.49 \\
        Conditional & Conv1d & 19,440 & 311,040 & 0.61 & 9.83 \\
        Residual & Conv1d & 14,520 & 217,800 & 0.46 & 6.91 \\
        Skip & Conv1d & 29,040 & 464,640 & 0.92 & 14.75\\
        \hline
        \textit{Post-processing Layers} & & \multicolumn{2}{r|}{} & \multicolumn{2}{c}{} \\
        Out & Conv1d & \multicolumn{2}{r|}{61,440} & \multicolumn{2}{r}{1.96} \\
        End & Conv1d & \multicolumn{2}{r|}{65,536} & \multicolumn{2}{r}{2.09} \\
        \hline
        \textit{Total} & & \multicolumn{2}{r|}{7,196,696} & \multicolumn{2}{r}{65.85} \\
    \end{tabular}
    \caption{A breakdown of the model's parameters and giga-operations required per second of synthesized audio output.}
    \label{tab:model_size}
\end{table*}

\section{Experiments}\label{sec:eval}

\subsection{Setup and Evaluation}

\subsubsection{Model}

\begin{figure}[!ht]
    \centering
    \includegraphics{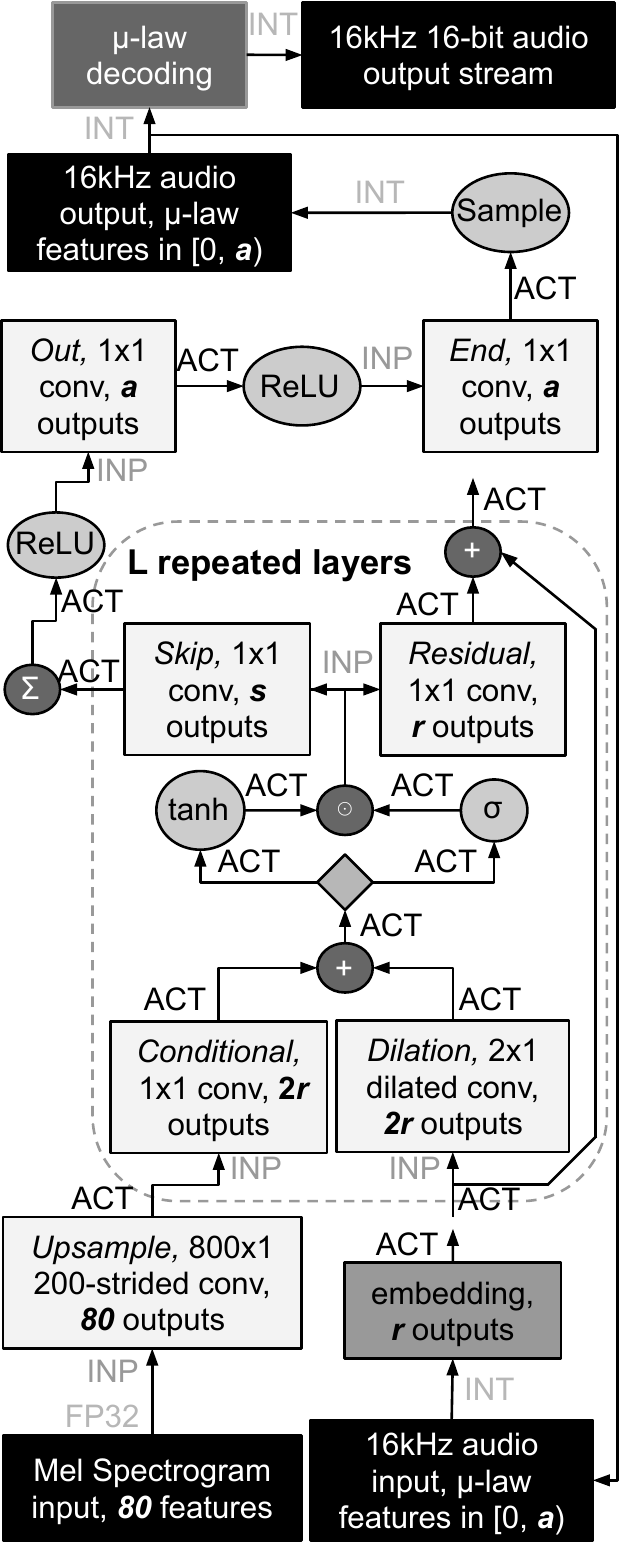}
    \caption{WaveNet model architecture. \textit{ACT} denotes the activation format before or after an operation. For the convolution layers, the input activations and parameters are both converted to the \textit{INP} format prior to use. \textit{INT} refers to a 64-bit integer data type. The diamond splits the input channels in half to make two outputs.}
    \label{fig:wavenet_bit_width}
\end{figure}

We use the WaveNet architecture as detailed in \figurename~\ref{fig:wavenet_bit_width}. This model uses mel spectrograms with 80 filter banks for the conditional features and upsamples them using a 1D transposed convolution with a kernel size of 800 and a stride of 200. The remainder of the architecture is parameterized by the number of channels in its convolutions: the number of skip, residual and audio channels; as well as by the number of repeated layers and dilation cycle parameter $D$. For all experiments we use a model with $\textbf{s} = 240$ skip channels, $\textbf{r} = 120$ residual channels, $\textbf{a} = 256$ audio channels, $\textbf{L} = 16$ repeated layers and $D = 8$.  This model has 7.2 million parameters, see Table~\ref{tab:model_size}. Whilst the model produces $256$ audio channels the final audio output has a bit depth of 16 as a $\mu$-law compounding transform is applied \cite{oord2016wavenet}.

\subsubsection{Sparsity}

The weights of the dilation, conditional, residual and skip convolutions in the repeated layers and the out post-processing convolution weights are pruned as combined these layers contain $95.6\%$ of the total operations; see \tablename~\ref{tab:model_size}. The feature upsample pre-processing weights are also pruned as this layer contains $71.1\%$ of the total number of parameters despite only requiring $1.25\%$ of the total operations. Biases for all pruned layers remain dense. Finally, the embedding pre-processing and end post-processing layers remain dense as, although they have comparatively few parameters and operations, pruning them has a disproportionate effect on the final quality.

We focus on two sparsity granularities as these are supported by the major deep learning frameworks including TensorFlow and PyTorch. The first granularity is layer-wise unstructured sparsity. We use iterative magnitude-based pruning where each layer is pruned independently up to a target compression ratio, defined as per Equation~\ref{eqn:CR}.
\begin{equation}
    \text{compression ratio} = \frac{\text{original size}}{\text{compressed size}}
    \label{eqn:CR}
\end{equation}

Each layer is pruned every 500 steps following the cubic pruning schedule found in TensorFlow \cite{zhu2017prune}. The second granularity is balanced sparsity where 2 out of every block of 4 values is pruned; denoted 2:4. For this technique we use NVIDIA's automatic sparsity library and follow their recommended approach: train a dense model, perform one-shot pruning to the target sparsity, and then repeat the training process starting from the pruned parameters.

\subsubsection{Quantization}

We use the 7 formats listed in \tablename~\ref{tab:precisions}. For our BFP16 implementation, we use a block size of 10 in the channel dimension. These formats are chosen as they are common amongst deep learning frameworks as well as current and future CPUs and neural network accelerators. PyTorch \cite{NEURIPS2019_9015} is used for INT8 and FP32. QPyTorch \cite{zhang2019qpytorch} is used to simulate bfloat16 \cite{intel2020bfloat16}, FP16.16, FP16.32, BFP16 \cite{song2017computation}, and TF32 \cite{nvidia2020a100}. Simulation is achieved by converting the input activations and parameters of the convolution layers to the target formats. The remaining operations may use a different format as detailed in \tablename~\ref{tab:precisions}.

\begin{table}[!htb]
    \centering
    {
        \begin{tabular}{llrrl}
            \textit{INP} Format & Abbrev. & E & M & \textit{ACT} \\
            \hline
            binary32 & FP32 & 8 & 23 & FP32 \\
            TensorFloat32 & TF32 & 8 & 10 & FP32 \\
            BrainFloat16 & bfloat16 & 8 & 7 & FP32 \\
            BlockFloat16 & BFP16 & 8 & 7 & FP32 \\
            binary16 & FP16.16 & 5 & 10 & FP16 \\
            binary16 & FP16.32 & 5 & 10 & FP32 \\
            Signed 8-bit int & INT8 & 0 & 8 & INT8 \\
        \end{tabular}
    }
    \caption{A summary of the numerical formats evaluated in this paper. The binary32 and binary16 formats are as defined in the IEEE 754-2008 standard. The E and M columns refer to the number of bits used for the exponent and mantissa respectively. The \textit{INP} and \textit{ACT} columns relate to the values used in \figurename~\ref{fig:wavenet_bit_width}.}
    \label{tab:precisions}
\end{table}

We use post-training quantization for all formats. We experimented with using QAT for some formats but found that it did not have a significant impact on the final model quality compared to PTQ despite being much harder to integrate into a training pipeline and increasing the barriers to deployment. Hence, we leave more detailed investigations regarding QAT to future work.

\subsubsection{Training}

We use the LJSpeech \cite{ljspeech17} dataset subsampled to 16kHz. Prior to starting this research, 100 samples were randomly selected from the dataset for use as a validation set and 100 samples were randomly selected from the dataset for use as a test set. All remaining samples are used as a training set. No data augmentation is applied.

All experiments use a batch size of 16 distributed across 1--4 GPUs as well as mixed precision training \cite{micikevicius2017mixed}. One element in the batch consists of a randomly selected 1 second segment from an audio clip in the training set, padding with zeros where necessary. We use the Adam optimizer \cite{kingma2014adam} with a fixed learning rate of $10^{-3}$, $\beta_1=0.9$, $\beta_2=0.999$, $\epsilon=10^{-8}$. 

\subsubsection{Evaluation}

We report both the teacher-forced cross-entropy validation and test loss, the Mean Opinion Score (MOS) and the compression ratio for all experiments. For the sparsity experiments we also report theoretical speedup. Each experiment is repeated 3 times with 3 different random seeds and the model with median validation loss is selected to compute and report all of these metrics. 

For each generated sample in the test set the MOS is computed by asking 30 independent Amazon Mechanical Turk workers to rate the sample's naturalness on a five point scale. The reported MOS is the mean of these scores and a 95\% confidence interval is computed using the t-distribution. 

The compression ratio is defined as per Equation~\ref{eqn:CR}. When referring to compression ratio or \textit{model} compression ratio the size is defined to be the total size of the model in bits. We also refer to the \textit{sparse layer} compression ratio where the size is defined to be the size in bits of only the layers that are being pruned. For sparse models, some layers remain dense so the model compression ratio is lower than the sparse layer compression ratio. The theoretical speedup is defined as the number of multiply-adds in the dense model divided by the number in the sparse model \cite{blalock2020state}.

Note that both the compression ratio and theoretical speedup only provide an upper bound on that achievable in deployment as, in practice, there will be extra overheads. For example, additional memory and hardware is required to store and use sparse parameters.

\subsection{Results}

\subsubsection{Sparsity}

\begin{table*}[ht]
    \centering
    \begin{tabular}{l|l|r|r|c|c|c|r}
        Granularity & Structure & \multicolumn{2}{c|}{Compression Ratio} & Val Loss & Test Loss & MOS & Theoretical Speedup \\
        & & Sparse Layer & Model & & & & \\
        \hline
        \hline
        \multicolumn{2}{l|}{Ground Truth (Human)} & \multicolumn{2}{r|}{-} & - & - & $4.072 \pm 0.028$ & - \\
        \hline
        \multicolumn{2}{l|}{Baseline (Dense)} & - & 1.00 & 2.189 & 2.182 & $3.976 \pm 0.027$ & 1.00 \\
        \hline
        Iterative & Unstructured & 2.00 & 1.97 & 2.179 & 2.173 & $3.966 \pm 0.029$ & 1.91 \\
        Iterative & Unstructured & 4.00 & 3.83 & 2.200 & 2.194 & $3.922 \pm 0.030$ & 3.51 \\
        Iterative & Unstructured & 8.00 & 7.23 & 2.236 & 2.231 & $3.877 \pm 0.032$ & 6.03 \\
        Iterative & Unstructured & 16.00 & 13.02 & 2.281 & 2.276 & $3.757 \pm 0.033$ & 9.41 \\
        Iterative & Unstructured & 32.00 & 21.73 & 2.357 & 2.402 & $3.616 \pm 0.038$ & 12.95 \\
        \hline
        One-shot & 2:4 & 2.00 & 1.97 & 2.195 & 2.191 & $3.881 \pm 0.027$ & 1.91 \\
    \end{tabular}
    \caption{Impact of varying sparsity techniques and compression ratios on model quality and the theoretical speedups.}
    \label{tab:sparse_mos}
\end{table*}

The results for the sparsity experiments that use the iterative unstructured and one-shot 2:4 magnitude-based pruning techniques are presented in \tablename~\ref{tab:sparse_mos}. 

\begin{figure}[ht]
    \centering
    \includegraphics{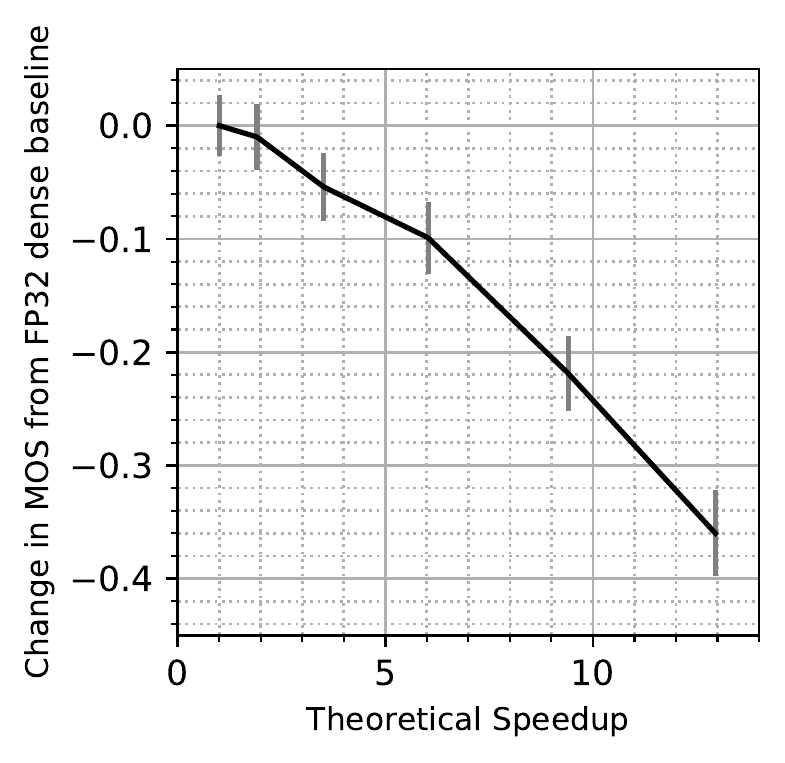}
    \caption{Trade-off between model quality (change in MOS from baseline) and the theoretical speedup. Error bars display 95\% confidence intervals using the t-distribution.}
    \label{fig:speedup_mos}
\end{figure}

Focusing on the iterative unstructured experiments, we see that the difference between the baseline model and the models that use a sparse layer compression ratio of 2 and 4 respectively is not statistically significant. This means a model compression ratio of up to 3.83 and a theoretical speedup of up to 3.51 can be achieved without a reduction in fidelity of the synthesized audio. The models with a sparse layer compression ratio greater than 4 do exhibit a significant degradation in audio fidelity. However, this reduction in fidelity does lead to a large increase in theoretical speedup. For example, the model that uses a compression ratio of 32 in its sparse layers---$4.6\%$ the total number of parameters and $7.7\%$ the total number of operations---has a potential theoretical speedup of 12.95 over the dense baseline model whilst achieving only a 9.1\% lower MOS. However, note that MOS is a subjective metric and therefore relative comparisons are difficult to interpret. \figurename~\ref{fig:speedup_mos} demonstrates this trade-off between model quality and the theoretical speedup that sparsity offers.

Looking at the one-shot 2:4 magnitude-based pruning result we see that it has a significantly lower MOS than both the baseline and iterative unstructured model with a sparse layer compression ratio of 2. Further, the MOS score is comparable to that of the iterative unstructured models that use a sparse layer compression ratio of 4 and 8. This suggests that the iterative unstructured magnitude-based pruning technique produces higher quality WaveNet models for a fixed compression ratio or of greater theoretical speedup for a fixed quality.

\subsubsection{Quantization}

\begin{table*}[ht]
    \centering
    \begin{tabular}{l|r|c|c|c}
        Format & Compression Ratio & Val Loss & Test Loss & MOS \\
        \hline
        \hline
        \multicolumn{2}{l|}{Ground Truth (Human)} & - & - & $4.072 \pm 0.028$ \\
        \hline
        FP32 & 1.00 & 2.189 & 2.182 & $3.976 \pm 0.027$ \\
        \hline
        TF32 & 1.68 & 2.189 & 2.182 & $3.972 \pm 0.029$ \\
        bfloat16 & 2.00 & 2.189 & 2.183 & $3.904 \pm 0.030$ \\
        BFP16 & 3.61 & 2.189 & 2.183 & $3.877 \pm 0.032$ \\
        FP16.16 & 2.00 & 2.189 & 2.182 & $3.911 \pm 0.029$ \\
        FP16.32 & 2.00 & 2.189 & 2.182 & $3.903 \pm 0.031$ \\
        INT8 & 4.00 & 2.442 & 2.435 & $3.877 \pm 0.032$ \\
    \end{tabular}
    \caption{Impact on model fidelity of post-training quantization.}
    \label{tab:quantization_mos}
\end{table*}

We present the quantization results in \tablename~\ref{tab:quantization_mos}. We are able to obtain an audio fidelity that matches that of the baseline FP32 model by using the TF32 precision. The other formats are all equivalent to each other but they are all significantly worse than TF32 and FP32. This could be due to the higher compression ratio with these formats compared to TF32 and FP32 causing a higher information loss in the model arithmetic that leads to a quality degradation.

\subsubsection{Sparsity and Quantization Combined}

\begin{table*}[ht]
    \centering
    \begin{tabular}{l||c|c|r|c||c|c|r|c}
        & \multicolumn{4}{c||}{Sparse Layer Compression Ratio 4} & \multicolumn{4}{c}{2:4} \\
        Format & Val Loss & Test Loss & CR & MOS & Val Loss & Test Loss & CR & MOS \\
        \hline
        \hline
        FP32 & 2.200 & 2.194 & 3.83 & $3.922 \pm 0.030$ & 2.195 & 2.191 & 1.97 & $3.881 \pm 0.027$ \\
        \hline
        TF32 & 2.200 & 2.194 & 6.44 & $3.951 \pm 0.031$ & 2.197 & 2.191 & 3.32 & $3.926 \pm 0.028$ \\
        bfloat16 & 2.201 & 2.195 & 7.65 & $3.894 \pm 0.030$ & 2.197 & 2.191 & 3.94 & $3.835 \pm 0.026$ \\
        BFP16 & 2.201 & 2.195 & 13.84 & $3.934 \pm 0.028$ & 2.197 & 2.191 & 7.13 & $3.862 \pm 0.029$ \\
        FP16.16 & 2.200 & 2.194 & 7.65 & $3.895 \pm 0.029$ & 2.197 & 2.191 & 3.94 & $3.830 \pm 0.031$ \\
        FP16.32 & 2.200 & 2.194 & 7.65 & $3.919 \pm 0.030$ & 2.197 & 2.191 & 3.94 & $3.889 \pm 0.029$ \\
        INT8 & 3.591 & 3.534 & 15.30 & $3.527 \pm 0.040$ & 2.635 & 2.629 & 7.88 & $3.836 \pm 0.030$ \\
    \end{tabular}
    \caption{Effect on model quality of post-training quantization of sparse models.}
    \label{tab:combination_mos}
\end{table*}

Finally, we consider applying both quantization and sparsity to our WaveNet model at the same time. Due to time constraints and the large number of runs required if running every combination of sparsity and precision previously considered, we choose to only investigate two different sparsity levels. We use a per layer compression ratio of 4 since this produced a model with a high compression ratio that maintains a similar quality to the dense baseline. We also investigate with the 2:4 sparsity pattern since we suspect that this will become a popular choice for sparsity with tools supported by NVIDIA for this sparsity level.

The results for these experiments are presented in Table~\ref{tab:combination_mos}. When looking at the models using the sparse layer compression ratio of 4, we find that using quantization does not provide a significant degradation in quality compared to our FP32 baseline in all cases besides INT8. Our best model with this sparsity pattern is the TF32 which achieve MOS comparable with a baseline FP32 dense model whilst having a compression ratio of over 6.

When looking at the models that use the 2:4 sparsity pattern, we find results similar to the ones obtained using the sparse layer compression ratio of 4. The quantized models all have comparable audio fidelity to the FP32 model with no significant change in MOS result, despite having a compression ratio of up to 7.88 in the case of INT8.

\section{Conclusions and Future Work}

In this work we have shown that model compression can be utilised to create a WaveNet vocoder with a high compression ratio whilst still being capable of near-human quality speech synthesis. Our BFP16 model with a sparse layer compression ratio of 4 synthesizes audio with a quality that is on par with a dense FP32 baseline whilst achieving a compression ratio of 13.84.

We hypothesise that higher compression ratios may be achieved by exploiting more extreme forms of quantization, such as INT4, INT2 and even binary models, although we suspect that the use of QAT will be vital to maintain acceptable audio fidelity in these cases.

We hope that our results encourage the use of model quantization and sparsity to realise the theoretical speedup afforded by them on next generation hardware accelerators to produce high quality text-to-speech systems, although we leave the specifics of such deployments to future work.

\bibliography{refs.bib}

\end{document}